\documentclass{article}

\usepackage[preprint]{neurips_2026}

\usepackage[utf8]{inputenc}
\usepackage[T1]{fontenc}
\usepackage{hyperref}
\usepackage{url}
\usepackage{xurl}
\usepackage{booktabs}
\usepackage{amsmath}
\usepackage{amsfonts}
\usepackage{amssymb}
\usepackage{nicefrac}
\usepackage{microtype}
\usepackage{xcolor}
\usepackage{graphicx}
\usepackage{array}
\usepackage{algorithm}
\usepackage{algpseudocode}

\title{Speculative Rollback Correction for Quality-Diverse Web Agent Imitation}

\author{
\small
\textbf{Longkun Hao}$^{1,*}$ \quad
\textbf{Hongyu Lin}$^{2,*}$ \quad
\textbf{Hao Li}$^{3}$ \quad
\textbf{Zhuowen Liu}$^{4}$ \quad
\textbf{Zhichao Yang}$^{1}$ \quad
\textbf{Haojie Hao}$^{1}$ \\
\textbf{Dongshuo Huang}$^{5}$ \quad
\textbf{Haitao Yang}$^{6}$ \quad
\textbf{Hongyu Ge}$^{7}$ \quad
\textbf{Mingjie Xie}$^{2}$ \quad
\textbf{Yanjun Wu}$^{2}$ \\
\textbf{Zihao Yin}$^{8}$ \quad
\textbf{Yan Bai}$^{8}$ \quad
\textbf{Yihang Lou}$^{8,\dagger}$ \\
\normalfont\scriptsize $^{1}$Beihang University \quad
$^{2}$Institute of Software, Chinese Academy of Sciences \\
\normalfont\scriptsize $^{3}$The Hong Kong University of Science and Technology \quad
$^{4}$The Chinese University of Hong Kong \\
\normalfont\scriptsize $^{5}$Northwestern Polytechnical University \quad
$^{6}$Tsinghua University \\
\normalfont\scriptsize $^{7}$The Hong Kong University of Science and Technology (Guangzhou) \quad
$^{8}$Peking University
}

\newcommand{\method}{Speculative Rollback Correction}
\newcommand{\src}{SRC}
\newcommand{\qd}{QD}
\newcommand{\codeurl}{https://github.com/LongkunHao/SRC_gui_agent}
\newcommand{\authornotes}{%
  \begingroup
  \renewcommand{\thefootnote}{\fnsymbol{footnote}}%
  \footnotetext[1]{Equal contribution.}%
  \footnotetext[2]{Corresponding author. Contact: \texttt{haolongkun@buaa.edu.cn}; \texttt{yihanglou@pku.edu.cn}.}%
  \endgroup
}
\newcommand{\E}{\mathbb{E}}
\newcommand{\Tau}{\mathcal{T}}
\newcommand{\Archive}{\mathcal{A}}
\newcommand{\Branch}{\mathcal{B}}
\newcommand{\Verifier}{\mathcal{V}}

\begin{document}

\maketitle
\authornotes

\begin{abstract}
Training interactive web agents through imitation learning from expert trajectories has emerged as a highly effective approach. However, Determining the optimal timing for expert intervention presents a critical challenge in this context. Delayed intervention often leads to the accumulation of early-stage errors, pushing the page state into an irrecoverable regime. Conversely, premature or excessive intervention causes the agent to become overly reliant on expert policies, trapping the model in local optima characterized by a single, rigid trajectory.
We propose \method{} (\src), a branch-level imitation framework for resettable agent environments. Instead of requesting teacher labels at every visited state or correcting only after a completed trajectory, \src{} uses fixed-horizon branch review: the student executes a short speculative segment before teacher review, and the teacher localizes the first harmful deviation only when local progress breaks. Rollback preserves useful prefixes, while successful rollouts are filtered by a hard verifier and retained in a lightweight quality-diversity archive. The resulting data supports next-action supervised fine-tuning on both localized corrections and verifier-passing trajectories. On WebArena-Infinity, \src{} collects 977 verifier-passing trajectories and 9,183 next-action examples; fixed-horizon review improves the recovery-versus-query tradeoff over step-level review while retaining verifier-passing solution variants. Code is available at \url{\codeurl}.

\end{abstract}

\section{Introduction}


Imitation learning is a natural starting point for training web and GUI agents: expert trajectories provide concrete demonstrations of how natural-language goals can be translated into browser or desktop actions. However, standard behavior cloning is brittle in long-horizon interactive environments. A student policy is not evaluated on the clean states visited by an expert, but on the states induced by its own previous actions. Once the student clicks the wrong element, enters a poor search query, opens an irrelevant page, or terminates prematurely, all subsequent observations may be generated by this mistake rather than by the expert path. This learner-induced distribution shift makes long-horizon GUI agents especially vulnerable to compounding error: one early action can move the agent into a state where the remaining expert trajectory is no longer meaningful.This failure mode is closely related to learning-to-search and exposure-bias phenomena in sequential prediction, where locally accurate one-step predictions may still induce poor long-horizon behavior once the learner conditions on its own past decisions \citep{daume2009search,chang2015learning,lamb2016professor,laskey2017dart}.

This problem becomes more severe in GUI and web tasks because success is rarely tied to a single canonical path.The same instruction may be solved through search, browsing, filters, keyboard shortcuts, menu navigation, or different valid subgoal orderings. However, single-path expert supervision can collapse these valid alternatives into one teacher-preferred behavior mode: a student action that deviates from the demonstrated path may be penalized even when it is the beginning of another successful solution. Conversely, accepting all student deviations preserves diversity but also admits loops, redundant exploration, and low-quality successful trajectories. The desired training signal is therefore not merely expert imitation or unfiltered self-exploration, but quality-constrained diversity: multiple verifier-passing solution paths that remain efficient and useful for training.

Interactive imitation learning and DAgger-style data aggregation study how to train policies on learner-induced states rather than only on expert demonstrations \citep{ross2011dagger}. Recent agent-training methods revisit this principle for LLM and tool agents. LEAP uses privileged teacher feedback to correct on-policy rollouts \citep{choudhury2025leap}, while On-policy Expert Corrections (OEC) start a trajectory with the student and switch to an expert partway through the rollout \citep{lauffer2025oec}. These methods establish an important point: agents should learn from states that the current student actually visits, not only from perfect expert demonstrations.

\begin{figure}[t]
  \centering
  \includegraphics[width=\linewidth]{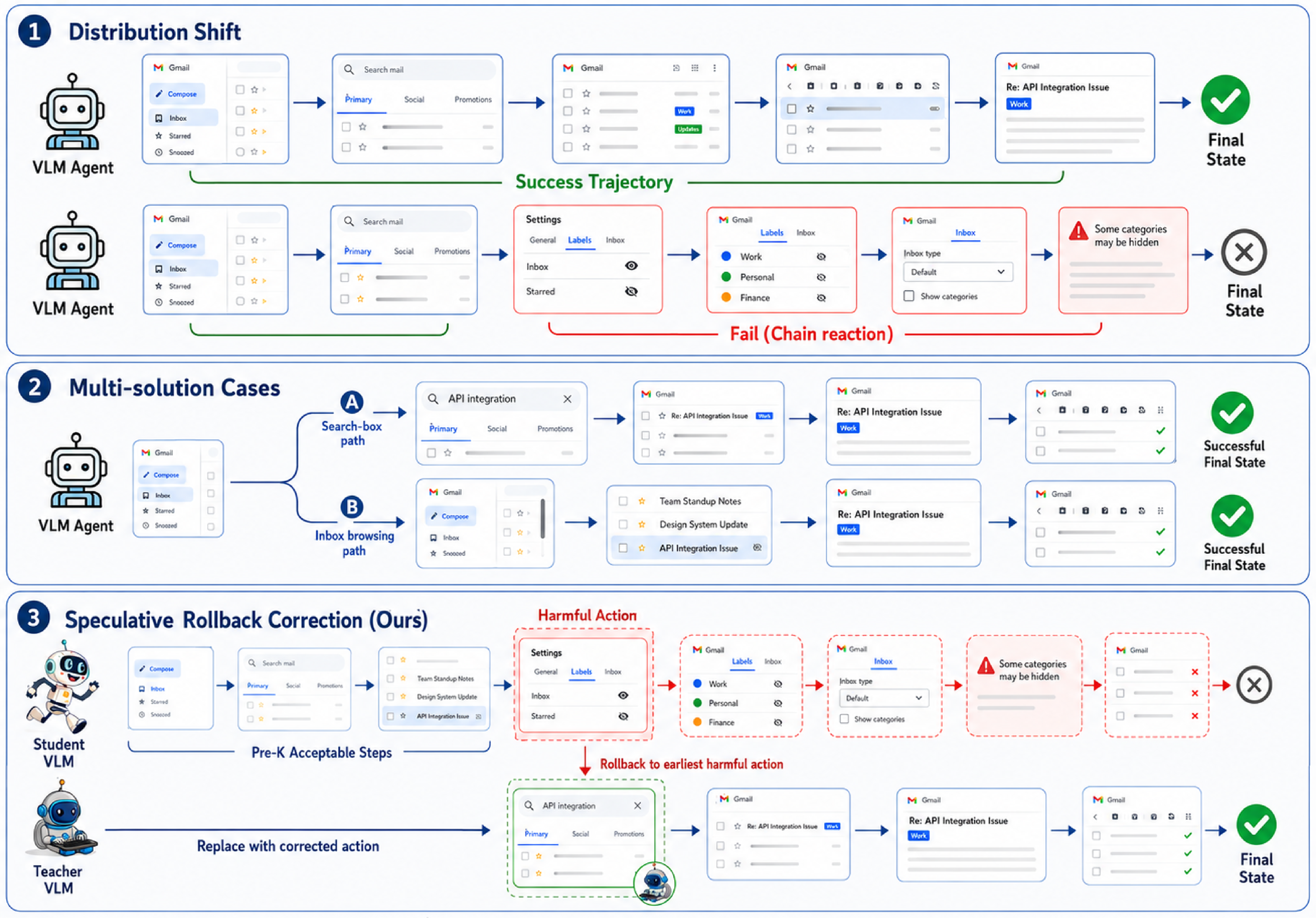}
  \caption{Two motivations for branch-level rollback correction. In long-horizon GUI tasks, one early mistake can push later observations off the successful trajectory, making post-hoc correction data-inefficient. At the same time, the same task can admit multiple verifier-passing paths, such as search-box retrieval and inbox browsing, so deviations from a teacher trajectory should not be treated as errors unless they harm local progress.}
  \label{fig:branch-rollback-motivation}
\end{figure}

However, existing online correction strategies leave open a key design question for visual, long-horizon GUI environments: at what granularity should teacher supervision be requested? Post-hoc correction can arrive after one early mistake has already corrupted many later states, while random takeover may interrupt useful exploration or miss the key failure point. Figure~\ref{fig:branch-rollback-motivation} illustrates this tension. In such environments, an effective correction mechanism should be delayed enough to preserve useful student exploration, localized enough to identify the first harmful action, and flexible enough to retain multiple verifier-passing solution paths.

We propose \method{} (\src), a training framework for resettable GUI environments. The core idea is to move teacher supervision from individual actions to short speculative branches. The student first executes a branch, and the teacher then judges whether the branch preserves local progress or where it first becomes harmful. When a harmful turn is found, rollback keeps the useful prefix and repairs only the suffix that caused the deviation. At episode end, a hard verifier decides whether the full trajectory succeeded, and a lightweight quality-diversity archive retains high-quality successful trajectories across different behavioral modes. This separates three roles that are often conflated: the teacher judges local progress, the verifier judges final success, and the archive decides which successful alternatives are worth training on.To our knowledge, SRC is the first framework to systematically adapt DAgger-style online expert correction for web and GUI agents, filling a long-standing gap in interactive imitation learning for realistic visual interaction scenarios.

This paper makes three contributions:
\begin{enumerate}
  \item We are the first to systematically implement DAgger-style online interactive expert correction for GUI and web agents. This allows the policy to learn from states encountered during real student rollouts, fundamentally mitigating the compounding error and state drift inherent in standard behavior cloning.

  \item We propose Speculative Rollback Correction (SRC), a branch-level training mechanism for interactive GUI tasks. By speculative short-branch rollout and precise minimal rollback, our method preserves valid student exploration and maintains behavioral quality-diversity, balancing both training stability and multi-solution learning.

  \item We conduct extensive evaluations on both web and desktop GUI benchmarks. SRC achieves consistent performance gains over baseline methods on long-horizon challenging tasks and shows strong cross-domain generalization.

\end{enumerate}
Our framework is model and modality-agnostic, and serves as a general training paradigm for long-horizon interactive agents to evolve from passive imitation to autonomous and reliable execution.
\section{Related Work}

\paragraph{Imitation learning and on-policy correction.}
Behavior cloning trains a policy on expert state-action pairs and is simple to scale, but its deployment-time state distribution can drift away from the expert distribution. This exposure-bias issue is well known in sequential prediction \citep{bengio2015scheduled} and was formalized for imitation learning by DAgger, which rolls out the learned policy and queries an expert on learner-induced states \citep{ross2011dagger}. Cost-aware interactive variants further show that the value of supervision depends on where mistakes occur in a trajectory \citep{ross2014aggrevate}. Earlier learning-to-search reductions such as SEARN and LOLS similarly frame structured prediction as sequential decision making and study how roll-ins, reference policies, and cost-to-go estimates shape the learned policy \citep{daume2009search,chang2015learning}. Recent work on language and web agents also moves beyond static expert demonstrations: LEAP uses privileged feedback to correct on-policy errors in LLM agents \citep{choudhury2025leap}, OEC generates partially on-policy trajectories by letting the student act first and then switching to an expert \citep{lauffer2025oec}, and Agent Q and WebRL collect interaction traces through search, self-critique, online curricula, or reinforcement learning \citep{putta2024agentq,qi2024webrl}. Related language-agent systems use browser interaction, tool calls, verbal reflection, or search over reasoning states to improve interaction-time behavior \citep{nakano2021webgpt,yao2023react,schick2023toolformer,shinn2023reflexion,yao2023tree,wang2023voyager}. \src{} follows this broad motivation but changes the correction unit: in resettable GUI environments, the student speculates for a short branch before the teacher localizes the earliest harmful action, preserving useful prefixes rather than requesting teacher labels before every action or relying only on full-trajectory post-hoc feedback.

\paragraph{Quality diversity and multi-solution trajectory data.}
In GUI-agent data collection, behavior descriptors can separate search-based, browsing-based, and correction-heavy solutions that all pass the same verifier. Quality-diversity methods seek a collection of high-performing solutions spread across behavior descriptors rather than a single optimum \citep{mouret2015mapelites,pugh2016qualitydiversity}. MAP-Elites is the canonical example: it partitions a descriptor space into niches and stores high-quality elites within each niche \citep{mouret2015mapelites}. We use this archive viewpoint for agent data collection rather than evolutionary search. Candidate solutions are complete web-agent trajectories, quality is enforced by a hard verifier and rule-based trajectory constraints, and descriptors organize different interaction modes for supervised training. This differs from keeping all successful rollouts in a replay buffer: the archive is intended to retain multiple verifier-passing ways of solving the same task while filtering loops, excessive detours, and correction-heavy trajectories.

\paragraph{Web and GUI agents.}
Early web-agent work studied executable browser tasks through text, DOM, accessibility trees, or filtered HTML, as in World of Bits, WebShop, Mind2Web, WebArena, and MiniWoB-style computer-use tasks \citep{shi2017world,yao2022webshop,deng2023mind2web,zhou2023webarena,kim2023rci}. More recent GUI agents increasingly use rendered screenshots or multimodal observations because graphical interfaces are designed for visual perception: VisualWebArena emphasizes visually grounded web tasks \citep{koh2024visualwebarena}, SeeAct studies GPT-4V-style web grounding with visual and HTML signals \citep{zheng2024seeact}, CogAgent demonstrates screenshot-grounded GUI navigation \citep{hong2024cogagent}, and OSWorld evaluates multimodal agents in real computer environments \citep{xie2024osworld}. Our work targets the long-horizon training problem shared by these environments rather than a specific observation modality: whether the agent receives text, DOM, accessibility, screenshot, or multimodal observations, learner-induced mistakes still compound and many tasks admit multiple verifier-passing interaction paths.

\section{Problem Setup}

We consider a set of GUI-agent tasks $\mathcal{X}$, including browser and desktop interaction environments. Each task $x \in \mathcal{X}$ defines a partially observed sequential decision problem with observation $o_t \in \mathcal{O}_x$, action $a_t \in \mathcal{A}$, transition kernel $P_x(o_{t+1}\mid o_t,a_t)$, and actor policy $\pi_\theta(a_t \mid h_t, x)$. The agent conditions on the interaction history
$
  h_t = (o_1,a_1,\ldots,o_t),
$
and a trajectory is $\tau=(o_1,a_1,\ldots,o_T,a_T)$. 

The environment provides a hard verifier 
$
  \Verifier(x,\tau) \in \{0,1\},
$
which checks whether the task has been completed. We write the verifier-passing set for task $x$ as
\[
  \Tau_x^+ = \{\tau:\Verifier(x,\tau)=1\}.
\]
We assume the training environment can be reset and replayed. This is the default for self-hosted benchmarks and generated browser environments used in this paper.

The learner is a student policy $\pi_\theta(a_t \mid h_t, x)$. Let $d_t^\pi(h\mid x)$ denote the distribution over histories at time $t$ induced by rolling out policy $\pi$ on task $x$, and let $d_t^E$ denote the history distribution in expert demonstrations. Standard behavior cloning minimizes the following objective:
\[
  \mathcal{L}_{\mathrm{BC}}(\theta)
  =
  \E_{x,t,h_t\sim d_t^E,\;a_t^E}
  [-\log \pi_\theta(a_t^E\mid h_t,x)],
\]
but deployment evaluates the same policy under its own induced history distribution $d_t^{\pi_\theta}$. After an early GUI mistake, $d_t^{\pi_\theta}$ can move away from $d_t^E$, so later observations may be generated by the student's error rather than by an expert path. A stronger teacher $T$ is available during data collection to assess these student-induced states and provide corrective supervision. At test time, the student acts without teacher access.

The second objective is not to recover a single expert mode or a single corrected path. In \src{}, one collection pass on task $x$ can yield a small set of candidate trajectories: the mainline follows the teacher-corrected path after rollback, while rejected student branches may remain as separate student-discovered leaves. We write this candidate set as
\[
  \mathcal{C}_x(\pi_\theta,T)
  =
  \{\tau_\ell:\ell\in\operatorname{Leaves}(\mathcal{G}_x)\},
\]
where $\mathcal{G}_x$ is the rollout tree induced by the student policy, teacher corrections, and reset-and-replay branching. The candidates in $\mathcal{C}_x$ may share prefixes but differ in whether later actions come from the student, the teacher-corrected mainline, or a mixture of both.

Many tasks have multiple verifier-passing trajectories, so $\Tau_x^+$ may contain qualitatively different interaction paths. We associate each candidate trajectory with a quality vector
\[
  q(\tau) = \bigl(|\tau|,\mathrm{repeat}(\tau),\mathrm{interventions}(\tau)\bigr)
\]
and a behavior descriptor $b(\tau)$ such as a path-length bucket, dominant action type, or teacher-intervention bucket. The desired data object is therefore an archive
\[
  \Archive_x \subseteq
  \{\tau\in\mathcal{C}_x(\pi_\theta,T)\cap\Tau_x^+:
  q(\tau)\leq q_{\max}\},
\]
where the quality constraint is applied componentwise. The archive should cover descriptor bins rather than only the shortest or most teacher-like trajectory.

Our goal is to collect training data that improves robust task success while preserving multiple valid solution modes. We therefore distinguish three judgments:
\begin{itemize}
  \item \textbf{Local progress:} whether a short branch continues to move toward the task without causing a meaningful recovery burden.
  \item \textbf{Final success:} whether the completed trajectory passes the hard verifier.
  \item \textbf{Trajectory quality:} whether a successful trajectory is efficient enough to train on, measured by length, repeated actions, loops, and teacher-intervention count.
\end{itemize}
The teacher handles local progress, the verifier handles final success, and the \qd{} archive handles high-quality multi-solution retention.

\section{Speculative Rollback Correction}

\subsection{Overview}

\begin{figure}[H]
  \centering
  \includegraphics[width=\linewidth]{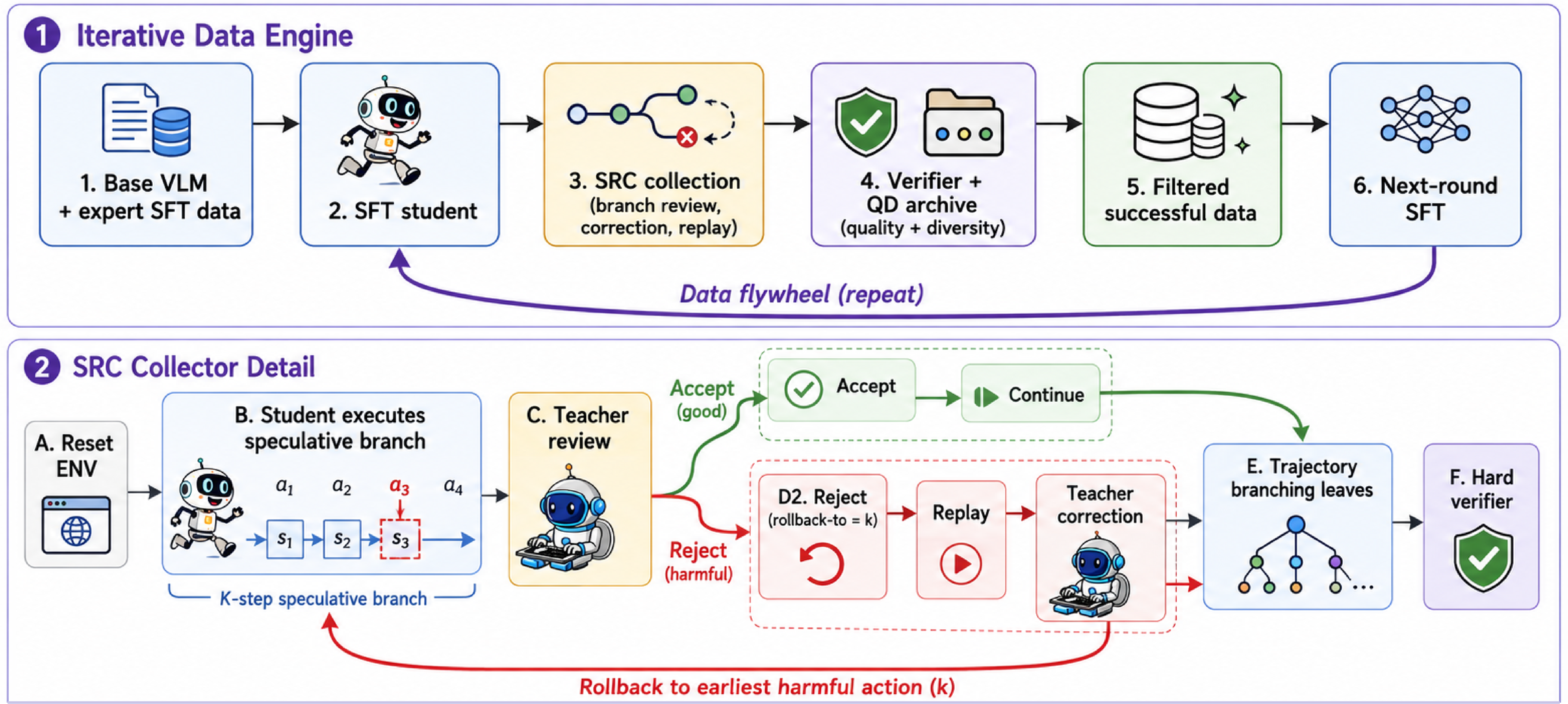}
  \caption{\method{} data engine and collector loop. The top row shows iterative data generation and SFT: a student collects \src{} trajectories, verifier and archive filtering retain successful data, and the next-round student is trained on the filtered mixture. The bottom row shows one \src{} collection step: a teacher first reviews a speculative branch as a rollback critic, and only after rollback does it generate a corrective action from the recovered observation.}
  \label{fig:overview}
\end{figure}

Figure~\ref{fig:overview} summarizes \method{} (\src). The method is a data-generation loop for resettable GUI environments. In each round, the current student collects trajectories under branch-level teacher review. Verifier-passing trajectories are filtered by a lightweight quality-diversity archive and then mixed with localized teacher corrections for the next round of supervised fine-tuning.

\src{} separates three decisions that are often conflated in expert-correction pipelines. A teacher reviewer judges \emph{local progress}: whether a short student branch remains recoverable and task-directed. A hard verifier judges \emph{final success}: whether a completed trajectory satisfies the task. A quality-diversity archive judges \emph{training value}: which successful trajectories are efficient and behaviorally distinct enough to keep. This separation lets the teacher correct harmful deviations without forcing all successful behavior onto one canonical path.By decoupling local progress judgment, global success verification, and diversity preservation, SRC overcomes the limitations of conventional step-level or post-hoc correction, forming an efficient and scalable interactive learning paradigm for real-world GUI environments.

\subsection{Fixed-Horizon Branch Review and Rollback Correction}

Rather than ask for a teacher label at every state, \src{} lets the student speculate for a short horizon. Starting from history $h_t$, the student executes a branch of at most $K$ actions,
\[
  \Branch_t =
  ((o_t,h_t,a_t), (o_{t+1},h_{t+1},a_{t+1}), \ldots,
  (o_{t+K-1},h_{t+K-1},a_{t+K-1})).
\]
The horizon $K$ is a review parameter, not a task constant: small $K$ approaches step-level correction, while larger $K$ reduces review frequency but can allow longer harmful suffixes before rollback.The branch horizon can be viewed as a supervision granularity parameter: at one extreme, step-level review resembles interactive imitation learning with frequent expert queries; at the other extreme, post-hoc review risks assigning correction only after many downstream states have already been contaminated by an early mistake \citep{ross2011dagger,ross2014aggrevate,daume2009search,chang2015learning}.

After the branch is executed, the teacher reviewer returns either \textsc{accept} or an earliest harmful index $j$. Acceptance means that the branch is locally progress-preserving, even if it follows a different valid route from the teacher's preferred solution. Rejection is reserved for objective failures such as entering the wrong object or page, creating a loop, drifting away from the instruction, making recovery substantially harder, or falsely terminating.

If the branch is accepted, all student actions are committed. If the branch is rejected at index $j$, \src{} keeps the useful prefix before $j$, restores the environment to the corresponding state, queries a teacher corrector on the recovered observation, executes the corrective action, and resumes student rollout from the corrected state. The first teacher query therefore localizes the failure; the second query is used only when rollback is needed and produces a next-action correction label.

\begin{algorithm}[t]
  \caption{Speculative Rollback Collector}
  \label{alg:src}
  \small
  \begin{algorithmic}[1]
    \Require task $x$, student $\pi_\theta$, reviewer $T_r$, corrector $T_c$, horizon $K$, verifier $\Verifier$
    \State Initialize environment, trajectory $\tau \leftarrow \emptyset$, corrections $\mathcal{D}_{\mathrm{corr}}\leftarrow\emptyset$
    \While{the episode has not terminated and budget remains}
      \State Roll out $\pi_\theta$ for a speculative branch $\Branch$ of at most $K$ actions
      \State Query $T_r$ to accept $\Branch$ or return earliest harmful index $j$
      \If{$T_r$ accepts}
        \State Commit $\Branch$ to $\tau$
      \Else
        \State Commit the prefix of $\Branch$ before $j$ to $\tau$
        \State Restore the environment to the rollback state
        \State Query $T_c$ for a corrective action $a_j^\star$ from the recovered observation
        \State Execute $a_j^\star$; add $(x,h_j,a_j^\star)$ to $\mathcal{D}_{\mathrm{corr}}$ and append the corrected step to $\tau$
      \EndIf
    \EndWhile
    \State \Return trajectory $\tau$, corrections $\mathcal{D}_{\mathrm{corr}}$, and verifier result $\Verifier(x,\tau)$
  \end{algorithmic}
\end{algorithm}

\subsection{Multi-Leaf Collection and Verifier-Constrained Archive}

The rollback path is the recovery path, but it need not be the only useful data. To preserve multiple valid solutions, \src{} can keep rejected student continuations as separate logical leaves when a fork budget permits. The mainline follows the teacher-corrected branch, while forked leaves are independently replayed, completed, and verified. Thus one collection pass can produce a small candidate set $\mathcal{C}_x(\pi_\theta,T)$ containing student, teacher-corrected, and hybrid trajectories. Final policy evaluation does not use this branching rule; it is only a data-collection mechanism.

The archive keeps successful candidates only when they pass the hard verifier and satisfy quality constraints:
\[
  \Archive_x \subseteq
  \{\tau \in \mathcal{C}_x(\pi_\theta,T):
    \Verifier(x,\tau)=1,\; q(\tau)\leq q_{\max}\}.
\]
This archive role is closer to quality-diversity data curation than to ordinary replay buffering: successful trajectories compete within coarse behavioral niches, so training data can preserve multiple verified solution modes instead of only the shortest or most teacher-like path \citep{mouret2015mapelites,pugh2016qualitydiversity,lehman2011abandoning,cully2015robots}.

The descriptor $b(\tau)$ assigns successful trajectories to coarse behavior bins, such as path-length bucket, dominant action type, and teacher-intervention count. Within each task and bin, the archive keeps a small number of high-quality trajectories. This archive is deliberately lightweight: its purpose is not to solve semantic mode discovery, but to prevent training data from collapsing to the shortest or most teacher-like trajectory while filtering verifier-passing traces that succeed through excessive detours or repeated recovery.



\subsection{Training Objective}

The final training set contains localized rollback corrections $\mathcal{D}_{\mathrm{corr}}$ and action labels extracted from archived successful trajectories $\mathcal{D}_{\mathrm{arc}}=\bigcup_x \Archive_x$. Let
\[
  \mathcal{D}_{\mathrm{arc}}^{\mathrm{act}}
  =
  \{(x,h,a):(x,\tau)\in\mathcal{D}_{\mathrm{arc}},\ (h,a)\in\tau\}
\]
be the next-action labels induced by archived trajectories, and let
$\mathcal{D}_{\mathrm{sft}}=\mathcal{D}_{\mathrm{corr}}\cup\mathcal{D}_{\mathrm{arc}}^{\mathrm{act}}$.
We use next-action supervised fine-tuning only, without reward modeling, DPO, or pairwise preference optimization:
\begin{align}
  \mathcal{L}(\theta)
  =
  \E_{(x,h,a)\sim \mathcal{D}_{\mathrm{sft}}}
  [-\log \pi_\theta(a \mid h,x)].
  \label{eq:loss}
\end{align}
Teacher corrections therefore teach recovery from student-induced states at localized rollback points, while archived trajectory labels train on complete verifier-passing trajectories, including alternative solutions discovered during collection. We do not tune a correction-versus-archive weighting hyperparameter; all experiments use the empirical per-example mixture produced by the collector and archive.

\section{Experiments}
\label{sec:experiment}
We evaluate whether \src{} improves GUI-agent training, transfers beyond the primary benchmark, trades off success and teacher cost through the review horizon $K$, and preserves quantitative evidence of trajectory diversity.

\subsection{Benchmarks, baselines, and metrics}
\label{sec:Experiment_Settings}
\paragraph{Benchmarks.}
WebArena-Infinity is the primary benchmark, with programmatic verifiers and official easy, medium, and hard task buckets \citep{zhou2026wainf}. We also evaluate on WebArena-Lite and an OSWorld subset covering \texttt{chrome}, \texttt{os}, and LibreOffice applications \citep{xie2024osworld}; the latter tests desktop transfer rather than full OSWorld leaderboard performance.

\paragraph{Baselines.}
We compare the base model, expert-trajectory SFT, the final model trained with filtered \src{} data, OEC-style random expert switch, and LEAP-style post-hoc privileged correction when available. We also report the SFT model while it is collecting \src{} data; this is a collection-time row, not a teacher-free test-time policy, because teacher corrections can be inserted during the rollout. Strong closed or external agents are included as references where numbers are available.

\paragraph{Metrics.}
The main metric is success rate (SR). We also report average episode steps and teacher queries per task. Teacher queries are data-collection cost: first-shot branch reviews plus second-shot corrective interventions after rollback. They are not test-time cost for teacher-free policy rows.

\paragraph{Implementation settings.}
The student is a vision-language GUI agent with a Qwen-style XML tool-call schema over a $1920\times1080$ viewport. Unless otherwise stated, \src{} uses $K=3$, max-forks 4, max-leaves 8, max-interventions 6, reset-and-replay rollback, and hard-verifier final success. The archive admits verifier-passing trajectories of length at most 60, with no more than four repeated actions and six teacher interventions. Final SFT uses the empirical mix of archived trajectory labels and rollback-correction labels.
\subsection{Main results}

Table~\ref{tab:main-results} combines the main results across WebArena-Infinity, WebArena-Lite, and the OSWorld subset using metrics shared by all benchmarks. We move the WebArena-Infinity difficulty breakdown to Appendix~\ref{tab:appendix-wai-breakdown} to avoid forcing WebArena-Lite and OSWorld into empty easy/medium/hard columns.

\begin{table}[t]
  \caption{Main results across benchmarks. Success rates are percentages. The ``collector'' rows report teacher-assisted data collection behavior and should not be interpreted as teacher-free test-time policies.}
  \label{tab:main-results}
  \centering
  \footnotesize
  \setlength{\tabcolsep}{3.2pt}
  \begin{tabular}{llccc}
    \toprule
    Benchmark & Method & SR & Avg. steps & Teacher q./task \\
    \midrule
    WebArena-Infinity & Base model & 15.8 & 29.83 & -- \\
    & Expert SFT & 25.3 & 12.06 & -- \\
    & SFT collector with \src{} & 42.5 & 29.58 & 12.01 \\
    & OEC-style random switch & 20.4 & 17.82 & 10.09 \\
    & LEAP-style post-hoc correction & 31.8 & 27.42 & 27.42 \\
    & Expert SFT + \src{} & 35.0 & 28.10 & -- \\
    & Kimi K2.5 & 39.7 & 30.20 & -- \\
    & Qwen3.5 Plus & 43.7 & 30.49 & -- \\
    \midrule
    WebArena-Lite & Base model & 15.2 & 18.48 & -- \\
    & Expert SFT & 20.5 & 15.34 & -- \\
    & SFT collector with \src{} & 36.0 & 31.24 & 11.90 \\
    & OEC-style random switch & 17.0 & 16.61 & 10.15 \\
    & LEAP-style post-hoc correction & 26.8 & 28.64 & 28.64 \\
    & Expert SFT + \src{} & 24.0 & 30.13 & -- \\
    & Kimi K2.5 & 29.0 & 32.05 & -- \\
    \midrule
    OSWorld subset$^\dagger$ & Base model & 23.47 & 12.88 & -- \\
    & Expert SFT & 27.22 & 12.81 & -- \\
    & SFT collector with \src{} & 43.27 & 16.82 & 5.29 \\
    & Expert SFT + \src{} & 40.15 & 11.86 & -- \\
    & EVA-CUA-8B & 46.48 & n.r. & -- \\
    \bottomrule
  \end{tabular}

\end{table}
The final teacher-free \src{} model improves over Expert SFT on all three settings: +9.7 SR on WebArena-Infinity, +3.5 on WebArena-Lite, and +12.9 on the OSWorld subset. The larger OSWorld gain suggests the rollback-collected data is not only fitting WebArena-Infinity trajectories. OEC underperforms Expert SFT despite teacher queries, indicating that random expert switching wastes corrections. LEAP improves over SFT on web tasks but uses roughly one teacher query per step during collection; the \src{} collector reaches higher SR with less collection cost on WebArena-Infinity and WebArena-Lite, while the final \src{} rows incur no teacher cost at test time. Closed agents remain stronger, but \src{} narrows the gap with an open student.
\subsection{Review-horizon ablation}

The fixed horizon $K$ controls when the teacher reviews a speculative branch. Panel (a) of Figure~\ref{fig:experiment-diagnostics} summarizes WebArena-Infinity runs over Gmail, Linear-account-settings, and GitLab Plan-and-Track; full numbers are in Appendix Tables~\ref{tab:k} and~\ref{tab:appendix-k-per-app}. We use $K=3$ because it gives the best aggregate SR, 51.9\%, with fewer teacher queries than $K=1$ (2529 vs. 3324). $K=7$ lowers cost further but drops to 50.6 SR. Thus step-level review overuses the teacher, while long branches delay rollback enough to lose some recoverable progress.

\subsection{Analysis studies}

\begin{figure}[t]
  \centering
  \includegraphics[width=0.8\linewidth]{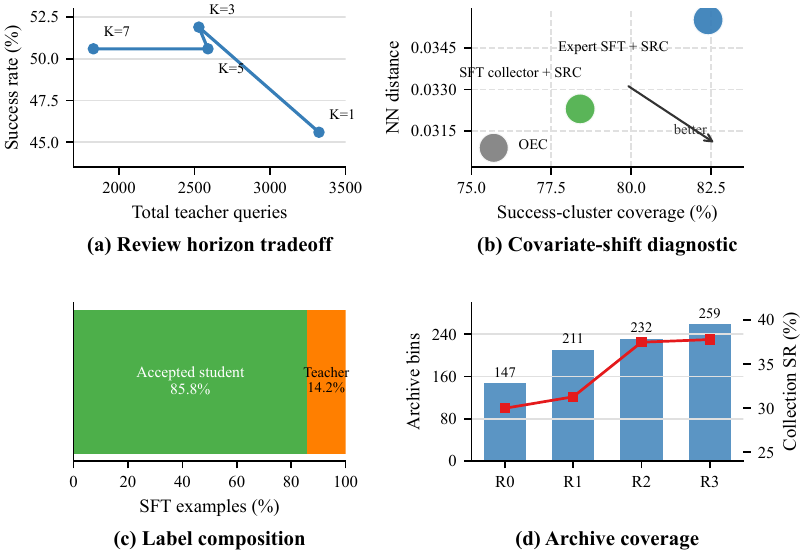}
  \caption{\src{} diagnostics. (a) Horizon $K$ trades success against teacher queries. (b) Covariate-shift diagnostic over success-cluster coverage and nearest-neighbor distance; circle area indicates effective modes. (c) Most SFT labels come from accepted branches. (d) Archive coverage grows across rounds, with later collection buying coverage through more intervention.}
  \label{fig:experiment-diagnostics}
\end{figure}

\paragraph{Covariate-shift diagnostics.}
We test whether \src{} changes visited states, not only final success labels, on a matched 100-task slice with expert, OEC, and \src{} trajectories. Screenshot states are embedded with a frozen Qwen3.5-9B VLM encoder and compared to the verifier-passing expert manifold by nearest-neighbor distance, state/cluster coverage, and effective covered modes. Panel (b) of Figure~\ref{fig:experiment-diagnostics} and Appendix Table~\ref{tab:covariate-shift} show that OEC reaches nearby successful regions but covers fewer success clusters. The \src{} collector and final policy cover more successful-state clusters while keeping nontrivial spread, supporting the claim that rollback correction reduces harmful drift without collapsing to one expert path.

\paragraph{Collection and archive diagnostics.}
Panels (c) and (d) summarize the final WebArena-Infinity export and archive growth; Appendix Tables~\ref{tab:appendix-data-export} and~\ref{tab:archive-coverage} give counts. The training set is not dominated by teacher actions: only 14.2\% of next-action examples come from rollback interventions, and content deduplication leaves 7,726 unique examples. Archive coverage expands from 147 to 259 retained descriptor bins, so the loop is not merely adding duplicate labels around one path. Because the R3 intervention rate is higher, we interpret this as a coverage--cost tradeoff rather than monotonic teacher-cost reduction. Appendix Figure~\ref{fig:casestudy} gives qualitative verifier-passing alternatives retained by the archive.

\section{Limitations}

\src{} assumes that the training environment can reset and replay branches with enough fidelity for rollback correction. This is realistic for self-hosted browser benchmarks and generated environments, but it limits the current method in non-resettable settings, irreversible workflows, or websites with hidden state changes.

The current implementation also uses a fixed review horizon $K$. This makes the ablation clean and keeps teacher-query accounting explicit, but a single horizon is unlikely to be optimal for all GUI tasks. Some subtasks require several low-level actions before meaningful progress can be judged, while others reach a decision point after one or two actions. A natural extension is adaptive branch review: the teacher first provides a global plan or subgoal decomposition, the student executes within the current subgoal, and teacher review is triggered when the student completes a subgoal or deviates from the plan. This would make the effective horizon depend on task structure rather than a fixed $K$.

\section{Conclusion}
\label{sec:conclusion}
This paper presents \method{}, a branch-level imitation framework for resettable GUI environments. Instead of asking for teacher labels at every step or correcting only after a full failed rollout, \src{} lets the student execute short speculative branches, rolls back to the earliest harmful action when needed, and keeps useful prefixes for next-action SFT. A verifier-constrained quality-diversity archive further preserves multiple successful solution modes.

The fixed-horizon ablation supports moderate branch review as a better recovery-versus-query tradeoff than step-level review, and the diagnostics show that \src{} data is not dominated by teacher interventions while archive coverage grows across collection rounds. Future work should relax the reset-and-replay assumption, extend Android evaluation, and use richer semantic trajectory descriptors.

\bibliographystyle{plainnat}
\bibliography{references}

@inproceedings{ross2011dagger,
  title = {A Reduction of Imitation Learning and Structured Prediction to No-Regret Online Learning},
  author = {Ross, Stephane and Gordon, Geoffrey and Bagnell, Drew},
  booktitle = {Proceedings of the Fourteenth International Conference on Artificial Intelligence and Statistics},
  series = {Proceedings of Machine Learning Research},
  volume = {15},
  pages = {627--635},
  year = {2011},
  publisher = {PMLR},
  url = {https://proceedings.mlr.press/v15/ross11a.html}
}

@inproceedings{bengio2015scheduled,
  title = {Scheduled Sampling for Sequence Prediction with Recurrent Neural Networks},
  author = {Bengio, Samy and Vinyals, Oriol and Jaitly, Navdeep and Shazeer, Noam},
  booktitle = {Advances in Neural Information Processing Systems},
  volume = {28},
  pages = {1171--1179},
  year = {2015},
  url = {https://papers.nips.cc/paper/5956-scheduled-sampling-for-sequence-prediction-with-recurrent-neural-networks}
}

@article{ross2014aggrevate,
  title = {Reinforcement and Imitation Learning via Interactive No-Regret Learning},
  author = {Ross, Stephane and Bagnell, J. Andrew},
  journal = {arXiv preprint arXiv:1406.5979},
  year = {2014},
  url = {https://arxiv.org/abs/1406.5979}
}

@inproceedings{choudhury2025leap,
  title = {Better than Your Teacher: {LLM} Agents that Learn from Privileged {AI} Feedback},
  author = {Choudhury, Sanjiban and Sodhi, Paloma},
  booktitle = {International Conference on Learning Representations},
  year = {2025},
  url = {https://openreview.net/forum?id=st7XqFgbAH}
}

@article{lauffer2025oec,
  title = {Imitation Learning for Multi-turn {LM} Agents via On-policy Expert Corrections},
  author = {Lauffer, Niklas and Deng, Xiang and Kundurthy, Srivatsa R. and Kenstler, Brad and Da, Jeff},
  journal = {arXiv preprint arXiv:2512.14895},
  year = {2025},
  url = {https://arxiv.org/abs/2512.14895}
}

@article{putta2024agentq,
  title = {Agent {Q}: Advanced Reasoning and Learning for Autonomous {AI} Agents},
  author = {Putta, Pranav and Mills, Edmund and Garg, Naman and Motwani, Sumeet and Finn, Chelsea and Garg, Divyansh and Rafailov, Rafael},
  journal = {arXiv preprint arXiv:2408.07199},
  year = {2024},
  url = {https://arxiv.org/abs/2408.07199}
}

@article{qi2024webrl,
  title = {{WebRL}: Training {LLM} Web Agents via Self-Evolving Online Curriculum Reinforcement Learning},
  author = {Qi, Zehan and Liu, Xiao and Iong, Iat Long and Lai, Hanyu and Sun, Xueqiao and Yang, Xinyue and Sun, Jiadai and Yang, Yu and Yao, Shuntian and Zhang, Tianjie and Xu, Wei and Tang, Jie and Dong, Yuxiao},
  journal = {arXiv preprint arXiv:2411.02337},
  year = {2024},
  url = {https://arxiv.org/abs/2411.02337}
}

@article{mouret2015mapelites,
  title = {Illuminating Search Spaces by Mapping Elites},
  author = {Mouret, Jean-Baptiste and Clune, Jeff},
  journal = {arXiv preprint arXiv:1504.04909},
  year = {2015},
  url = {https://arxiv.org/abs/1504.04909}
}

@article{pugh2016qualitydiversity,
  title = {Quality Diversity: A New Frontier for Evolutionary Computation},
  author = {Pugh, Justin K. and Soros, Lisa B. and Stanley, Kenneth O.},
  journal = {Frontiers in Robotics and AI},
  volume = {3},
  pages = {40},
  year = {2016},
  doi = {10.3389/frobt.2016.00040},
  url = {https://www.frontiersin.org/articles/10.3389/frobt.2016.00040/full}
}

@inproceedings{yao2022webshop,
  title = {{WebShop}: Towards Scalable Real-World Web Interaction with Grounded Language Agents},
  author = {Yao, Shunyu and Chen, Howard and Yang, John and Narasimhan, Karthik},
  booktitle = {Advances in Neural Information Processing Systems},
  volume = {35},
  year = {2022},
  url = {https://papers.nips.cc/paper_files/paper/2022/hash/82ad13ec01f9fe44c01cb91814fd7b8c-Abstract-Conference.html}
}

@inproceedings{deng2023mind2web,
  title = {{Mind2Web}: Towards a Generalist Agent for the Web},
  author = {Deng, Xiang and Gu, Yu and Zheng, Boyuan and Chen, Shijie and Stevens, Sam and Wang, Boshi and Sun, Huan and Su, Yu},
  booktitle = {Advances in Neural Information Processing Systems},
  volume = {36},
  year = {2023},
  url = {https://papers.nips.cc/paper_files/paper/2023/hash/5950bf290a1570ea401bf98882128160-Abstract-Datasets_and_Benchmarks.html}
}

@article{zhou2023webarena,
  title = {{WebArena}: A Realistic Web Environment for Building Autonomous Agents},
  author = {Zhou, Shuyan and Xu, Frank F. and Zhu, Hao and Zhou, Xuhui and Lo, Robert and Sridhar, Abishek and Cheng, Xianyi and Ou, Tianyue and Bisk, Yonatan and Fried, Daniel and Alon, Uri and Neubig, Graham},
  journal = {arXiv preprint arXiv:2307.13854},
  year = {2023},
  url = {https://arxiv.org/abs/2307.13854}
}

@inproceedings{koh2024visualwebarena,
  title = {{VisualWebArena}: Evaluating Multimodal Agents on Realistic Visual Web Tasks},
  author = {Koh, Jing Yu and Lo, Robert and Jang, Lawrence and Duvvur, Vikram and Lim, Ming Chong and Huang, Po-Yu and Neubig, Graham and Zhou, Shuyan and Salakhutdinov, Ruslan and Fried, Daniel},
  booktitle = {Proceedings of the 62nd Annual Meeting of the Association for Computational Linguistics (Volume 1: Long Papers)},
  pages = {881--905},
  year = {2024},
  publisher = {Association for Computational Linguistics},
  doi = {10.18653/v1/2024.acl-long.50},
  url = {https://aclanthology.org/2024.acl-long.50/}
}

@inproceedings{zheng2024seeact,
  title = {{GPT}-4{V}(ision) is a Generalist Web Agent, if Grounded},
  author = {Zheng, Boyuan and Gou, Boyu and Kil, Jihyung and Sun, Huan and Su, Yu},
  booktitle = {Proceedings of the 41st International Conference on Machine Learning},
  series = {Proceedings of Machine Learning Research},
  volume = {235},
  pages = {61349--61385},
  year = {2024},
  publisher = {PMLR},
  url = {https://proceedings.mlr.press/v235/zheng24e.html}
}

@inproceedings{hong2024cogagent,
  title = {{CogAgent}: A Visual Language Model for {GUI} Agents},
  author = {Hong, Wenyi and Wang, Weihan and Lv, Qingsong and Xu, Jiazheng and Yu, Wenmeng and Ji, Junhui and Wang, Yan and Wang, Zihan and Dong, Yuxiao and Ding, Ming and Tang, Jie},
  booktitle = {Proceedings of the IEEE/CVF Conference on Computer Vision and Pattern Recognition},
  pages = {14281--14290},
  year = {2024},
  url = {https://openaccess.thecvf.com/content/CVPR2024/html/Hong_CogAgent_A_Visual_Language_Model_for_GUI_Agents_CVPR_2024_paper.html}
}

@inproceedings{xie2024osworld,
  title = {{OSWorld}: Benchmarking Multimodal Agents for Open-Ended Tasks in Real Computer Environments},
  author = {Xie, Tianbao and Zhang, Danyang and Chen, Jixuan and Li, Xiaochuan and Zhao, Siheng and Cao, Ruisheng and Hua, Toh Jing and Cheng, Zhoujun and Shin, Dongchan and Lei, Fangyu and Liu, Yitao and Xu, Yiheng and Zhou, Shuyan and Savarese, Silvio and Xiong, Caiming and Zhong, Victor and Yu, Tao},
  booktitle = {Advances in Neural Information Processing Systems},
  volume = {37},
  year = {2024},
  url = {https://papers.nips.cc/paper_files/paper/2024/hash/5d413e48f84dc61244b6be550f1cd8f5-Abstract-Datasets_and_Benchmarks_Track.html}
}

@misc{zhou2026wainf,
  title = {{WebArena-Infinity}: Generating Browser Environments with Verifiable Tasks at Scale},
  author = {Zhou, Shuyan},
  year = {2026},
  month = {March},
  howpublished = {\url{https://webarena.dev/webarena-infinity/}},
  note = {GitHub repository: \url{https://github.com/web-arena-x/webarena-infinity}}
}

@inproceedings{kim2023rci,
  title = {Language Models Can Solve Computer Tasks},
  author = {Kim, Geunwoo and Baldi, Pierre and McAleer, Stephen},
  booktitle = {Advances in Neural Information Processing Systems},
  volume = {36},
  year = {2023},
  url = {https://proceedings.neurips.cc/paper_files/paper/2023/hash/7cc1005ec73cfbaac9fa21192b622507-Abstract-Conference.html}
}

@article{daume2009search,
  title = {Search-based Structured Prediction},
  author = {Daum{\'e} III, Hal and Langford, John and Marcu, Daniel},
  journal = {Machine Learning},
  volume = {75},
  number = {3},
  pages = {297--325},
  year = {2009}
}

@inproceedings{chang2015learning,
  title = {Learning to Search Better than Your Teacher},
  author = {Chang, Kai-Wei and Krishnamurthy, Akshay and Agarwal, Alekh and Daum{\'e} III, Hal and Langford, John},
  booktitle = {Proceedings of the 32nd International Conference on Machine Learning},
  series = {Proceedings of Machine Learning Research},
  volume = {37},
  pages = {2058--2066},
  year = {2015}
}

@inproceedings{laskey2017dart,
  title = {{DART}: Noise Injection for Robust Imitation Learning},
  author = {Laskey, Michael and Lee, Jonathan and Fox, Roy and Dragan, Anca and Goldberg, Ken},
  booktitle = {Proceedings of the 1st Annual Conference on Robot Learning},
  series = {Proceedings of Machine Learning Research},
  volume = {78},
  pages = {143--156},
  year = {2017}
}

@inproceedings{lamb2016professor,
  title = {Professor Forcing: A New Algorithm for Training Recurrent Networks},
  author = {Lamb, Alex M. and Goyal, Anirudh and Zhang, Ying and Zhang, Saizheng and Courville, Aaron C. and Bengio, Yoshua},
  booktitle = {Advances in Neural Information Processing Systems},
  volume = {29},
  year = {2016}
}

@article{nakano2021webgpt,
  title = {{WebGPT}: Browser-assisted Question-Answering with Human Feedback},
  author = {Nakano, Reiichiro and Hilton, Jacob and Balaji, Suchir and Wu, Jeff and Ouyang, Long and Kim, Christina and Hesse, Christopher and Jain, Shantanu and Kosaraju, Vineet and Saunders, William and Jiang, Xu and Cobbe, Karl and Eloundou, Tyna and Krueger, Gretchen and Button, Kevin and Knight, Matthew and Chess, Benjamin and Schulman, John},
  journal = {arXiv preprint arXiv:2112.09332},
  year = {2021},
  url = {https://arxiv.org/abs/2112.09332}
}

@inproceedings{yao2023react,
  title = {{ReAct}: Synergizing Reasoning and Acting in Language Models},
  author = {Yao, Shunyu and Zhao, Jeffrey and Yu, Dian and Du, Nan and Shafran, Izhak and Narasimhan, Karthik and Cao, Yuan},
  booktitle = {International Conference on Learning Representations},
  year = {2023},
  url = {https://openreview.net/forum?id=WE_vluYUL-X}
}

@inproceedings{schick2023toolformer,
  title = {Toolformer: Language Models Can Teach Themselves to Use Tools},
  author = {Schick, Timo and Dwivedi-Yu, Jane and Dess{\`i}, Roberto and Raileanu, Roberta and Lomeli, Maria and Hambro, Eric and Zettlemoyer, Luke and Cancedda, Nicola and Scialom, Thomas},
  booktitle = {Advances in Neural Information Processing Systems},
  volume = {36},
  year = {2023}
}

@inproceedings{shinn2023reflexion,
  title = {Reflexion: Language Agents with Verbal Reinforcement Learning},
  author = {Shinn, Noah and Cassano, Federico and Berman, Edward and Gopinath, Ashwin and Narasimhan, Karthik and Yao, Shunyu},
  booktitle = {Advances in Neural Information Processing Systems},
  volume = {36},
  year = {2023}
}

@inproceedings{yao2023tree,
  title = {Tree of Thoughts: Deliberate Problem Solving with Large Language Models},
  author = {Yao, Shunyu and Yu, Dian and Zhao, Jeffrey and Shafran, Izhak and Griffiths, Thomas L. and Cao, Yuan and Narasimhan, Karthik},
  booktitle = {Advances in Neural Information Processing Systems},
  volume = {36},
  year = {2023}
}

@article{wang2023voyager,
  title = {Voyager: An Open-Ended Embodied Agent with Large Language Models},
  author = {Wang, Guanzhi and Xie, Yuqi and Jiang, Yunfan and Mandlekar, Ajay and Xiao, Chaowei and Zhu, Yuke and Fan, Linxi and Anandkumar, Anima},
  journal = {arXiv preprint arXiv:2305.16291},
  year = {2023},
  url = {https://arxiv.org/abs/2305.16291}
}

@inproceedings{shi2017world,
  title = {World of Bits: An Open-Domain Platform for Web-Based Agents},
  author = {Shi, Tianlin and Karpathy, Andrej and Fan, Linxi and Hernandez, Jonathan and Liang, Percy},
  booktitle = {Proceedings of the 34th International Conference on Machine Learning},
  series = {Proceedings of Machine Learning Research},
  volume = {70},
  pages = {3135--3144},
  year = {2017}
}

@article{lehman2011abandoning,
  title = {Abandoning Objectives: Evolution through the Search for Novelty Alone},
  author = {Lehman, Joel and Stanley, Kenneth O.},
  journal = {Evolutionary Computation},
  volume = {19},
  number = {2},
  pages = {189--223},
  year = {2011}
}

@article{cully2015robots,
  title = {Robots that Can Adapt Like Animals},
  author = {Cully, Antoine and Clune, Jeff and Tarapore, Danesh and Mouret, Jean-Baptiste},
  journal = {Nature},
  volume = {521},
  number = {7553},
  pages = {503--507},
  year = {2015},
  doi = {10.1038/nature14422}
}

\appendix

\section{SRC Collection Details}

\subsection{Teacher Review and Correction Interface}

For each branch, the teacher reviewer receives the task instruction, recent pre-branch context, the pre-action observation for each student action, the executed action dictionaries, the student rationale when available, and the post-branch observation. If the branch proposes \texttt{terminate}, hard-verifier feedback is also passed to the reviewer. The reviewer returns one of the following:
\begin{itemize}
  \item \textsc{accept}: the branch is locally progress-preserving, even if it differs from the teacher's preferred path.
  \item \textsc{rollback}: the first harmful step index and a concise reason.
\end{itemize}
The teacher should prefer the earliest rollback point that preserves all useful preceding progress. The index is prefix-preserving: \texttt{rollback\_to}$=k$ keeps branch steps $0,\ldots,k-1$ and discards $k,\ldots,K-1$. The teacher should not roll back solely for stylistic differences, different but reasonable search terms, or alternate ordering of independent subtasks.

The corrective intervention is generated by a separate teacher query after reset-and-replay restores the rollback state. The rollback reason is included in the correction prompt so the teacher avoids repeating the rejected action. Correction responses use the same XML tool-call schema as the student. A correction may contain multiple low-level tool calls, but we count it as one teacher intervention for budgeting and query accounting.

\subsection{Teacher Prompts}

We use the following first-shot rollback-review prompt and corrective-intervention prompt.

\paragraph{Rollback review prompt.}
\begin{figure}
    \centering
    \includegraphics[width=1\linewidth]{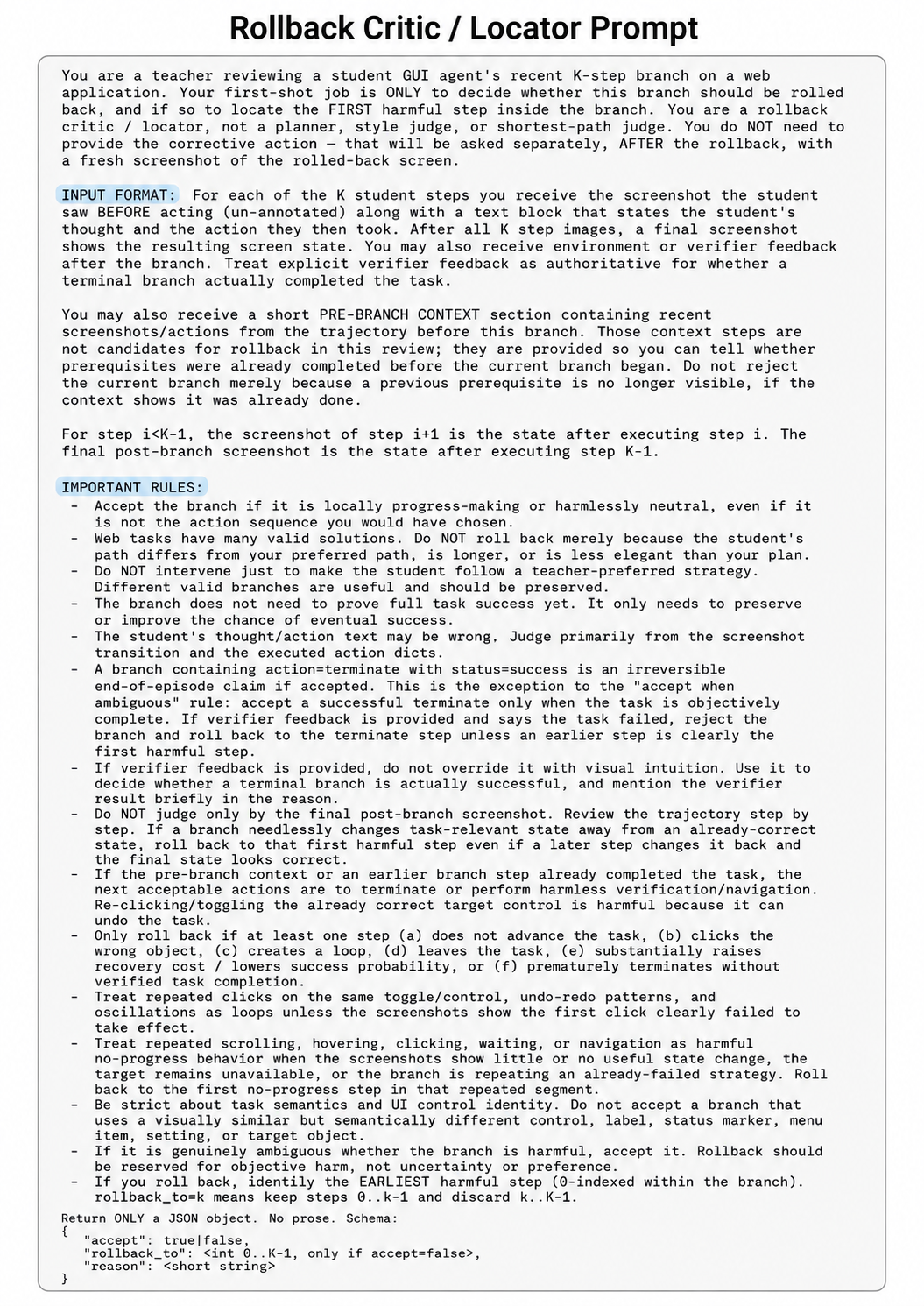}
    \caption{Rollback-review}
    \label{fig:Rollback-review}
\end{figure}
\begin{figure}
    \centering
    \includegraphics[width=1\linewidth]{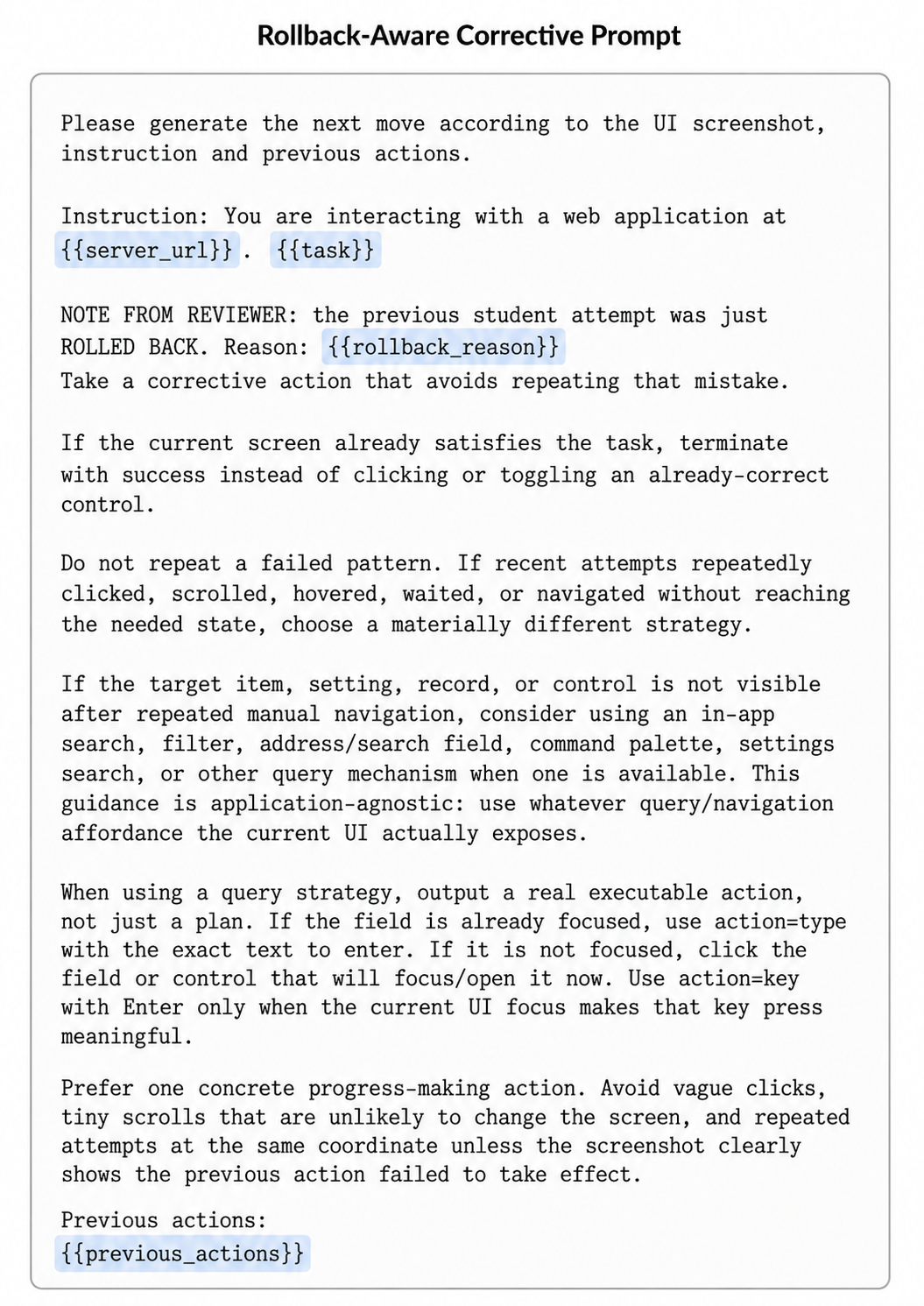}
    \caption{Corrective-intervention prompt}
    \label{fig:teacher_correct}
\end{figure}
\subsection{Reset-and-Replay}

When rollback is required, the collector restores the accepted prefix by resetting the environment, reopening the application, and replaying all actions before the rollback index. Replay skips \texttt{terminate} actions but preserves hover and mouse-move actions, because some GUI controls are only exposed after hover. Teacher corrections are inserted into the subsequent history as ordinary action records, which keeps the prompt state aligned with the environment state.

\subsection{Archive Filters and Descriptors}

The archive admits only verifier-passing trajectories satisfying the rule-based constraints
\[
  |\tau|\leq 60,\quad
  \mathrm{repeat}(\tau)\leq 4,\quad
  \mathrm{interventions}(\tau)\leq 6.
\]
Each admitted trajectory is assigned to a descriptor bin based on path-length bucket, dominant action type, and teacher-intervention count bucket. The path-length buckets are short ($\leq 5$ steps), medium ($\leq 12$), long ($\leq 25$), and extra-long ($>25$). Dominant action type is chosen from click, type, scroll, key, other, or none. Intervention count is bucketed as 0, 1, 2, or 3+. Each task/bin keeps at most three trajectories, preferring higher quality when available, otherwise shorter trajectories and fewer teacher interventions.

\subsection{Multi-Solution Case Studies}

\begin{figure}[p]
  \centering
  \includegraphics[width=\linewidth]{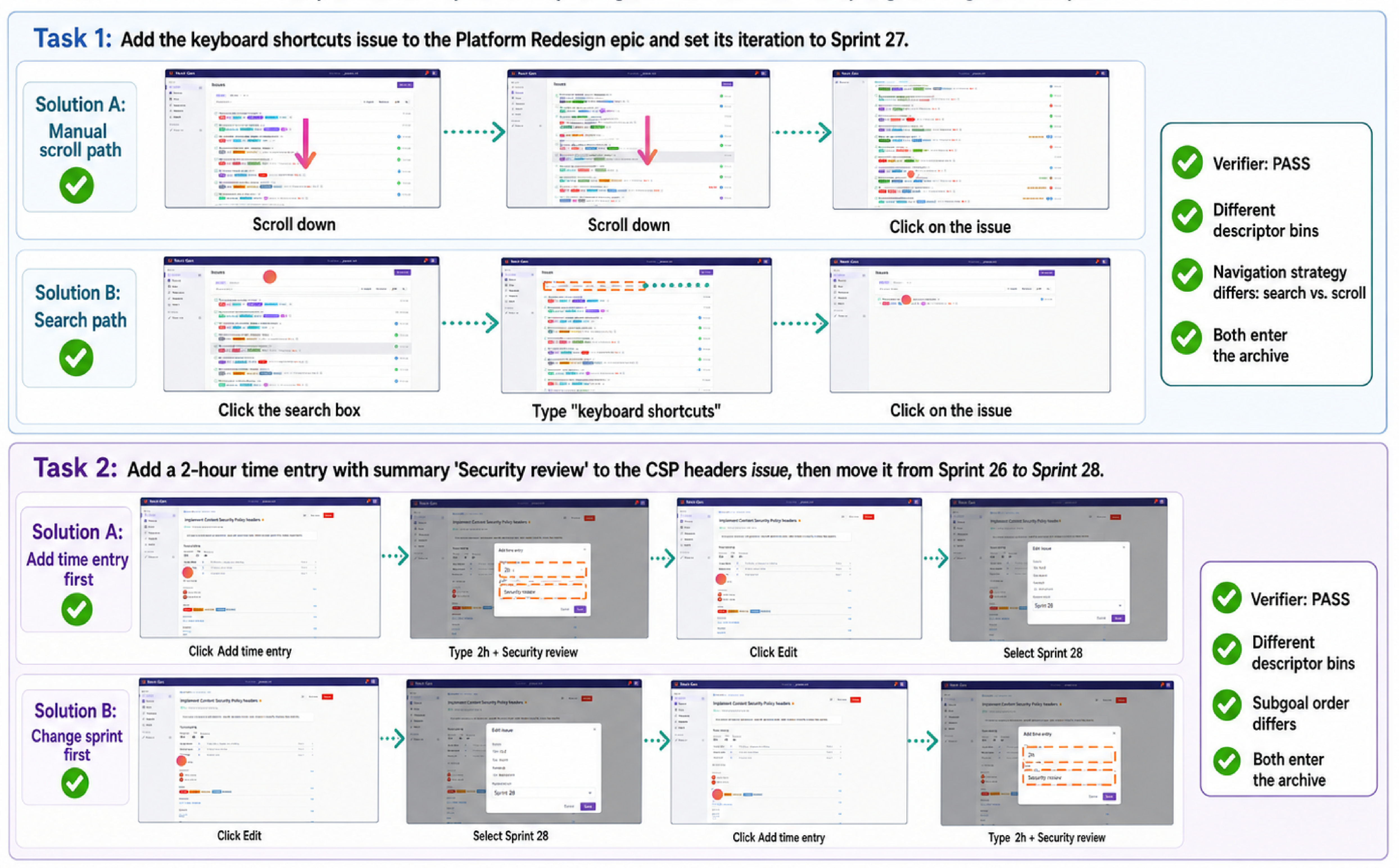}
  \caption{Case study 1. The retained trajectory illustrates a verifier-passing solution variant in which locally useful student progress is preserved instead of being overwritten by a single teacher-preferred path.}
  \label{fig:casestudy}
\end{figure}

Figures~\ref{fig:casestudy} show qualitative case studies retained by the archive. Across these examples, the useful variation is not coordinate-level jitter: accepted branches can preserve alternative navigation choices, rollback can remove only the first harmful suffix, and the corrected trajectory can still complete the task through a valid non-expert route. These cases support the archive design, which keeps verifier-passing solution variants rather than collapsing every task to one canonical demonstration.

\section{Additional Experimental Details}

\subsection{WebArena-Infinity Data Export}

The current WebArena-Infinity data export contains 977 verifier-passing QD trajectories and 9,183 next-action SFT examples. Among the SFT examples, 7,882 come from accepted student-branch actions and 1,301 come from teacher interventions after rollback. The trajectory difficulty distribution is 279 easy, 244 medium, and 454 hard. We also run basic export validation before training: JSON files are readable, referenced image paths exist, each example has system, user, and assistant messages, and each assistant message follows the expected \texttt{Action:} plus \texttt{<tool\_call>} format. Deduplicating by message content and image content leaves 7,726 unique examples; the remaining 15.9\% are duplicate shared prefixes from multi-leaf rollouts.

\begin{table}[h]
  \caption{WebArena-Infinity data export summary.}
  \label{tab:appendix-data-export}
  \centering
  \small
  \begin{tabular}{lr}
    \toprule
    Statistic & Value \\
    \midrule
    Verifier-passing QD trajectories & 977 \\
    Next-action SFT examples & 9,183 \\
    Student-branch examples & 7,882 \\
    Teacher-intervention examples & 1,301 \\
    Easy, medium, hard trajectories & 279, 244, 454 \\
    Unique examples after content deduplication & 7,726 \\
    Duplicate shared-prefix examples & 15.9\% \\
    \bottomrule
  \end{tabular}
\end{table}

\subsection{Covariate-Shift Diagnostics}

Table~\ref{tab:covariate-shift} gives the detailed numbers behind panel (b) of Figure~\ref{fig:experiment-diagnostics}. The \src{} collector row uses the 94 selected tasks with available collection trajectories. Distance is cosine nearest-neighbor distance to verifier-passing expert states in Qwen3.5-9B VLM state-embedding space. Coverage measures the fraction of successful expert states or success clusters reached within the automatically calibrated radius.

\begin{table}[h]
  \caption{Covariate-shift diagnostics on a 100-task WebArena-Infinity slice with matched expert, OEC, and final \src{} evaluation trajectories.}
  \label{tab:covariate-shift}
  \centering
  \footnotesize
  \setlength{\tabcolsep}{3.0pt}
  \resizebox{\linewidth}{!}{
  \begin{tabular}{lcccccc}
    \toprule
    Method & SR & States & NN dist. $\downarrow$ & State cov. $\uparrow$ & Cluster cov. $\uparrow$ & Eff. modes $\uparrow$ \\
    \midrule
    Expert trajectories & 100.0 & 1224 & 0.000 & 100.0 & 100.0 & 52.9 \\
    OEC random switch & 0.0 & 683 & 0.031 & 95.3 & 75.7 & 39.0 \\
    SFT collector with \src{} & 92.6 & 643 & 0.032 & 96.0 & 78.4 & 41.6 \\
    Expert SFT + \src{} & 74.0 & 725 & 0.035 & 97.0 & 82.4 & 41.8 \\
    \bottomrule
  \end{tabular}}
\end{table}

\subsection{Archive Coverage Across Rounds}

Table~\ref{tab:archive-coverage} gives the round-level counts behind panel (d) of Figure~\ref{fig:experiment-diagnostics}. Archive bins count unique descriptor bins retained after verifier and quality filtering.

\begin{table}[h]
  \caption{Archive coverage across collection rounds on WebArena-Infinity.}
  \label{tab:archive-coverage}
  \centering
  \small
  \setlength{\tabcolsep}{4pt}
  \begin{tabular}{lccc}
    \toprule
    Collection round & Collection SR & Interv./task & Archive bins \\
    \midrule
    R0: base model & 30.0 & 9.52 & 147 \\
    R1: expert SFT & 31.3 & 10.74 & 211 \\
    R2: SFT + \src{} data & 37.5 & 8.54 & 232 \\
    R3: recollect with R2 model & 37.8 & 34.84 & 259 \\
    \bottomrule
  \end{tabular}
\end{table}

\subsection{WebArena-Infinity Reference Runs}

Table~\ref{tab:appendix-sft-collecting} gives the full logged statistics for the SFT model while collecting \src{} data. Since usage and image-token accounting were not logged, we report teacher-query counts and a text-only token proxy.

\begin{table}[h]
  \caption{Full statistics for the SFT model collecting \src{} data on WebArena-Infinity.}
  \label{tab:appendix-sft-collecting}
  \centering
  \small
  \begin{tabular}{lr}
    \toprule
    Statistic & Value \\
    \midrule
    WAI-E SR & 87.1\% \\
    WAI-M SR & 67.1\% \\
    WAI-H SR & 26.2\% \\
    Overall SR on completed tasks & 43.2\% \\
    Average steps & 9.60 \\
    Teacher queries per completed task & 27.72 \\
    Total teacher queries & 41,252 \\
    Teacher review queries & 23,700 \\
    Teacher intervention queries & 17,552 \\
    Teacher text-token proxy & 89.122M text-only \\
    Logged task status & 1,488 completed; 693 timeout; 8 stuck; 124 missing \\
    \bottomrule
  \end{tabular}
\end{table}

Table~\ref{tab:appendix-closed-agent} reports the complete closed-agent references on the WebArena-Infinity real-task split. The split has 260 easy, 260 medium, and 1,100 hard tasks.

\begin{table}[h]
  \caption{WebArena-Infinity difficulty breakdown for the main-result rows.}
  \label{tab:appendix-wai-breakdown}
  \centering
  \small
  \begin{tabular}{lccc}
    \toprule
    Method & Easy SR & Medium SR & Hard SR \\
    \midrule
    Base model & 42.3 & 21.7 & 8.1 \\
    Expert SFT & 66.9 & 42.3 & 11.5 \\
    SFT collector with \src{} & 87.1 & 67.1 & 26.2 \\
    OEC-style random switch & 57.3 & 43.1 & 6.3 \\
    Expert SFT + \src{} & 67.3 & 56.2 & 22.3 \\
    Kimi K2.5 & 69.2 & 57.3 & 28.6 \\
    Qwen3.5 Plus & 77.3 & 56.5 & 32.7 \\
    \bottomrule
  \end{tabular}
\end{table}

\begin{table}[h]
  \caption{Closed-agent references on WebArena-Infinity real tasks. Average steps are computed over successful trajectories.}
  \label{tab:appendix-closed-agent}
  \centering
  \small
  \begin{tabular}{llcc}
    \toprule
    Model & Difficulty & Success & Avg. steps \\
    \midrule
    Kimi K2.5 & Easy & 69.23\% & 5.32 \\
    Kimi K2.5 & Medium & 57.31\% & 9.07 \\
    Kimi K2.5 & Hard & 28.55\% & 14.26 \\
    Kimi K2.5 & All & 39.69\% & 10.55 \\
    Qwen3.5 Plus & Easy & 77.31\% & 7.00 \\
    Qwen3.5 Plus & Medium & 56.54\% & 9.91 \\
    Qwen3.5 Plus & Hard & 32.73\% & 14.33 \\
    Qwen3.5 Plus & All & 43.70\% & 11.33 \\
    \bottomrule
  \end{tabular}
\end{table}

\subsection{Per-App Branch-Horizon Ablation}

Table~\ref{tab:k} gives the aggregate branch-horizon ablation behind panel (a) of Figure~\ref{fig:experiment-diagnostics}. Review queries are first-shot branch reviews; interventions are second-shot corrective teacher queries after rollback. Total teacher queries sum the two.

\begin{table}[h]
  \caption{Ablation over fixed speculative review horizon on WebArena-Infinity tasks aggregated across Gmail, Linear-account-settings, and GitLab Plan-and-Track.}
  \label{tab:k}
  \centering
  \footnotesize
  \setlength{\tabcolsep}{2.6pt}
  \resizebox{\linewidth}{!}{
  \begin{tabular}{cccccccccc}
    \toprule
    $K$ & SR & Easy SR & Medium SR & Hard SR & Review q. & Interv. & Total q. & Avg. rollback & Accept rate \\
    \midrule
    1 & 45.6 & 81.7 & 75.0 & 26.0 & 2603 & 721 & 3324 & 1.00 & \textbf{63.7} \\
    3 & \textbf{51.9} & \textbf{91.7} & \textbf{76.7} & 32.5 & 1744 & 785 & 2529 & 2.20 & 37.1 \\
    5 & 50.6 & 83.3 & 71.7 & \textbf{34.5} & 1708 & 881 & 2589 & 2.73 & 23.0 \\
    7 & 50.6 & 81.7 & \textbf{76.7} & 33.5 & \textbf{1227} & \textbf{604} & \textbf{1831} & 3.43 & 20.9 \\
    \bottomrule
  \end{tabular}}
\end{table}

Table~\ref{tab:appendix-k-per-app} expands the branch-horizon ablation by application.

\begin{table}[h]
  \caption{Per-app fixed-horizon ablation on WebArena-Infinity.}
  \label{tab:appendix-k-per-app}
  \centering
  \scriptsize
  \setlength{\tabcolsep}{2.5pt}
  \resizebox{\linewidth}{!}{
  \begin{tabular}{lccccccccc}
    \toprule
    App & $K$ & SR & Easy SR & Medium SR & Hard SR & Review q. & Interv. & Total q. & Accept \\
    \midrule
    GitLab Plan-and-Track & 1 & 22.1 & 60.0 & 50.0 & 9.0 & 413 & 40 & 453 & 85.2 \\
    GitLab Plan-and-Track & 3 & 28.6 & 80.0 & 65.0 & 11.0 & 353 & 116 & 469 & 51.8 \\
    GitLab Plan-and-Track & 5 & 28.6 & 70.0 & 45.0 & 17.0 & 537 & 237 & 774 & 29.8 \\
    GitLab Plan-and-Track & 7 & 27.9 & 65.0 & 60.0 & 14.0 & 272 & 99 & 371 & 30.1 \\
    Gmail & 1 & 65.0 & 85.0 & 75.0 & 35.0 & 1046 & 388 & 1434 & 52.4 \\
    Gmail & 3 & 65.0 & 95.0 & 65.0 & 35.0 & 373 & 192 & 565 & 28.4 \\
    Gmail & 5 & 60.0 & 80.0 & 70.0 & 30.0 & 273 & 145 & 418 & 20.9 \\
    Gmail & 7 & 60.0 & 80.0 & 70.0 & 30.0 & 252 & 120 & 372 & 19.8 \\
    Linear Account Settings & 1 & 63.3 & 100.0 & 100.0 & 45.0 & 1144 & 293 & 1437 & 66.2 \\
    Linear Account Settings & 3 & 72.5 & 100.0 & 100.0 & 58.8 & 1018 & 477 & 1495 & 35.2 \\
    Linear Account Settings & 5 & 71.7 & 100.0 & 100.0 & 57.5 & 898 & 499 & 1397 & 19.5 \\
    Linear Account Settings & 7 & 72.5 & 100.0 & 100.0 & 58.8 & 703 & 385 & 1088 & 17.8 \\
    \bottomrule
  \end{tabular}}
\end{table}

\end{document}